\documentclass{article}

\usepackage{microtype}
\usepackage{graphicx}
\usepackage{subfigure}
\usepackage{booktabs} % for professional tables
\usepackage{xcolor}
\usepackage{amsmath}
\usepackage{amssymb}
\usepackage{url}
\usepackage{subfigure}
\usepackage{placeins}
\usepackage{caption}
\usepackage{flushend}
\usepackage{lipsum}
\usepackage{float}
\usepackage{amsfonts}
\usepackage{makecell}

\usepackage{hyperref}

\usepackage[accepted]{icml2018}

\icmltitlerunning{prDeep: Robust Phase Retrieval with a Flexible Deep Network}

\begin{document}

\twocolumn[
\icmltitle{prDeep: Robust Phase Retrieval with a Flexible Deep Network}

% It is OKAY to include author information, even for blind
% submissions: the style file will automatically remove it for you
% unless you've provided the [accepted] option to the icml2018
% package.

% List of affiliations: The first argument should be a (short)
% identifier you will use later to specify author affiliations
% Academic affiliations should list Department, University, City, Region, Country
% Industry affiliations should list Company, City, Region, Country

% You can specify symbols, otherwise they are numbered in order.
% Ideally, you should not use this facility. Affiliations will be numbered
% in order of appearance and this is the preferred way.
%\icmlsetsymbol{equal}{*}

\begin{icmlauthorlist}
\icmlauthor{Christopher A. Metzler}{rice}
\icmlauthor{Philip Schniter}{OSU}
\icmlauthor{Ashok Veeraraghavan}{rice}
\icmlauthor{Richard G. Baraniuk}{rice}
\end{icmlauthorlist}

\icmlaffiliation{rice}{Department of Electrical and Computer Engineering, Rice University, Houston, TX}
\icmlaffiliation{OSU}{The Ohio State University, Columbus, OH}

\icmlcorrespondingauthor{Chris Metzler}{chris.metzler@rice.edu}

% You may provide any keywords that you
% find helpful for describing your paper; these are used to populate
% the "keywords" metadata in the PDF but will not be shown in the document
\icmlkeywords{Machine Learning, ICML}

\vskip 0.3in
]

% this must go after the closing bracket ] following \twocolumn[ ...

% This command actually creates the footnote in the first column
% listing the affiliations and the copyright notice.
% The command takes one argument, which is text to display at the start of the footnote.
% The \icmlEqualContribution command is standard text for equal contribution.
% Remove it (just {}) if you do not need this facility.

\printAffiliationsAndNotice{}  % leave blank if no need to mention equal contribution
%\printAffiliationsAndNotice{\icmlEqualContribution} % otherwise use the standard text.

\begin{abstract}
Phase retrieval algorithms have become an important component in many modern computational imaging systems.  
For instance, in the context of ptychography and speckle correlation imaging, they enable imaging past the diffraction limit and through scattering media, respectively. 
Unfortunately, traditional phase retrieval algorithms struggle in the presence of noise. 
Progress has been made recently on more robust algorithms using signal priors, but at the expense of limiting the range of supported measurement models (e.g., to Gaussian or coded diffraction patterns).  
In this work we leverage the regularization-by-denoising framework and a convolutional neural network denoiser to create {\em prDeep}, a new phase retrieval algorithm that is both robust and broadly applicable.   
We test and validate prDeep in simulation to demonstrate that it is robust to noise and can handle a variety of system models.
\end{abstract}

\section{Introduction}
%What is phase retrieval?
The PR problem manifests when one wants to recover the input from only the amplitude, or intensity, of the output of a linear system. Mathematically, PR refers to the problem of recovering a vectorized signal $x \in \mathbb{R}^n~\text{or}~\mathbb{C}^n$ from measurements $y$ of the form 
\begin{align}
y=|\mathbf{A}x|+w,
\end{align}
where the measurement matrix $\mathbf{A}$ represents the forward operator of the system and $w$ represents noise.

%What is it used for?
PR shows up in many imaging applications including microscopy \cite{zheng2013wide}, crystallography \cite{millane1990phase}, astronomical imaging \cite{fienup1987phase}, and inverse scattering \cite{katz2014non, MetzlerICCP}, to name just a few.

%How has it been done historically?
PR algorithms were first developed in the early 1970s and have been continuously studied by the optics community since then \cite{gerchberg1972practical,fienup1978reconstruction, HIO, griffin1984signal,ERHIO, OSS}. %bauschke2003hybrid, ERHIO,NRHIO,OSS}. 
More recently, PR has been taken up by the optimization community. This has produced a number of algorithms with theoretical, if not always practical, benefits \cite{candes2013phaselift,candes2015phase,goldstein2016phasemax,bahmani2017phase}. For a benchmark study of over a dozen popular PR algorithms, see PhasePack \cite{chandra2017phasepack}.

Following the popularity of compressive sensing, numerous algorithms were developed that use prior information, oftentimes sparsity, to improve PR reconstructions and potentially enable {\em compressive} PR \cite{moravec2007compressive, prGAMP}. In general, these methods do not improve reconstructions when the signal of interest is dense.

%How is it done now?
In the last few years, three methods, SPAR \cite{SPAR1,SPAR2}, BM3D-prGAMP \cite{bm3dprgamp}, and Plug-and-Play ADMM \cite{P&PADMM,Proximal}, have been developed to solve the PR problem using natural-image priors, which make their reconstructions more robust to noise. These methods all apply a natural-image prior via the BM3D image-denoising algorithm \cite{BM3D}.

%What's wrong with how it's done now?
Unfortunately, two of these three methods, SPAR and BM3D-prGAMP, are restricted to i.i.d Gaussian or coded diffraction measurements \cite{candesCodedDiffraction}, which prevents their use in many practical applications.
All three methods are computationally demanding.

%How do I plan to improve it?
In this work, we make two technical contributions. First, we show how the Regularization by Denoising (RED) \cite{RED} framework can be adapted to solve the PR problem. We call this adaption prRED. Because it sets up a general optimization problem, rather than using a specific algorithm, prRED is flexible and can handle a wide variety of measurements including, critically, Fourier measurements.

Second, we show how prRED can utilize convolutional neural networks by incorporating the DnCNN neural network \cite{DnCNN}. We call this combination of RED and DnCNN applied to PR prDeep.

%How significant are my changes.
prDeep offers excellent performance with reasonable run times. In Section \ref{sec:Results}, we apply prDeep to simulated data and show that it compares favorably to existing algorithms with respect to computation time and robustness to noise.

\section{Related Work}
%Can provide overview of PR algorithms in the intro
%\subsection{PR}
%Alternating projection,
%Gradient descent,
%Convex surrogate,
%Bayesian

Our work fits into the recent trend of using advanced signal priors to solve inverse problems in imaging. While priors like sparsity, smoothness, and structured sparsity have been studied for some time, we focus here on plug-and-play priors and deep-learning priors, which together represent the state-of-the-art for a range of imaging recovery tasks.

\subsection{Plug-and-Play Regularization for Linear Inverse Problems}\label{ssec:pandp}
%Image denoisers serve as the foundation for plug and play regularizatoin.
Image denoising is arguably the most fundamental problem in image processing and, as such, it has been studied extensively. Today there exist hundreds of denoising algorithms that model and exploit the complex structure of natural images in order to remove additive noise.

Earlier this decade, researchers realized they could leverage these highly developed denoising algorithms to act as regularizers in order to solve other linear inverse-problems, such as deblurring, superresolution, and compressed sensing \cite{BM3DasReg}.
This technique was eventually coined {\em plug-and-play} regularization \cite{P&PADMM}, with the idea being that one could ``plug in'' a denoiser to impose a specific prior on the inverse problem. A number of techniques have been developed that use this idea to solve various linear inverse problems \cite{BM3DasReg,P&PADMM,FlexISP,Proximal, DAMP,DVAMP,RED}.

Most of these works implicitly assume that the denoiser is a proximal mapping for some cost function $R(x)$, i.e.,
\begin{align}
D(z)=\text{prox}_R(z)\triangleq\arg\min_x\frac{1}{2}\|z-x\|^2_2+R(x),
\end{align}
where $R(x)$ penalizes image hypotheses that are unnatural. From the maximum a posteriori (MAP) Bayesian perspective, $R(x)$ is the negative log-prior for the natural-image $x$, and $z$ is a Gaussian-noise corrupted measurement of $x$.

With this interpretation, a variety of algorithms, such as (Plug-and-Play) ADMM \cite{P&PADMM,FlexISP,Proximal} or (Denoising-based) AMP \cite{DAMP,DVAMP}, can be used to recover $x$ from the linear measurements $y$ by solving the optimization problem
\begin{align}
\arg\min_x\frac{1}{2}\|y-\mathbf{A}x\|^2_2+ R(x),
\end{align}
where again $R(x)$ is an implicit cost function associated with the denoiser.
%\textcolor{red}{I took out $\lambda$ because it does not appear in other expressions, but we can add it back if needed.}

Because the priors associated with advanced denoisers like BM3D accurately model the distribution of natural images, these methods have offered state-of-the-art recovery accuracy in many of the tasks to which they have been applied. 

%By contrast, the regularization by denoising (RED) framework, which we describe in section \ref{ssec:RED}, creates an explicit cost function with the denoiser. This has a number of important benefits.

\subsection{Plug-and-Play Regularization for PR}\label{ssec:P&PforPR}

Following their success on linear inverse problems, plug-and-play priors were applied to the PR problem as well. We are aware of three prior works that take this approach.

The first, SPAR \cite{SPAR1, SPAR2}, uses alternating minimization to compute the MAP estimate of $x$ by using BM3D and a Poisson noise model; $y^2=\text{Poisson}(|z|^2)$, with $z\triangleq\mathbf{A}x$.
So far, the algorithm has only been succesfully applied to coded-diffraction pattern measurements.

%There seem to be a lot of TUT PR algorithms. Figure out which ones are relevant https://www.osapublishing.org/DirectPDFAccess/0D201DE5-AD26-0165-1BB94C73FEE0750E_225946/josaa-29-1-105.pdf?da=1&id=225946&seq=0&mobile=no
The second, BM3D-prGAMP \cite{bm3dprgamp}, uses the generalized approximate message passing (AMP) framework \cite{AMP,GAMP,prGAMP,DAMP} to compute the minimum mean squared error (MMSE) estimate of $x$ using BM3D and a Rician channel model: $y=|z+w|$ with $w\sim CN(0,\sigma_w^2)$. 
It requires the elements of $\mathbf{A}$ to be nearly i.i.d.~Gaussian, which can be approximated using coded-diffraction-pattern measurements. 

% Most recently, the authors of ProxImaL\footnote{ProxImaL is a language and compiler designed specifically for image optimization problems. It can set up and solve imaging problem using a number of algorithms including Pock-Chambolle algorithm, ADMM, linearized ADMM, and half-quadratic splitting.} \cite{Proximal} used Plug and Play ADMM to form a MAP estimate of $x$ using BM3D (or a TV prior) and a Gaussian measurement-noise model. The authors did so by solving the following optimization problem.
% \begin{align}
% \arg\min_x\frac{1}{2}\|y-|\mathbf{A}x|\|^2_2+R(x).
% \end{align}
% This data fidelity term, $\|y-|\mathbf{A}x|\|^2_2$, corresponds to the negative log-likelihood of $z$ under the model $y=|z|+w$ where $w\sim N(0,\sigma_w^2)$. Unlike the aforementioned methods, ADMM supports generic measurement matrices, including Fourier measurements.

Most recently, the authors of ProxImaL%\footnote{ProxImaL is a language and compiler designed specifically for image optimization problems.} % It can set up and solve imaging problems using a number of algorithms, including the Chambolle-Pock algorithm, ADMM, linearized ADMM, and half-quadratic splitting.} 
\cite{Proximal} used Plug-and-Play ADMM to estimate $x$ by solving the optimization problem 
%using BM3D (or a TV prior)  a Gaussian measurement-noise model. The authors did so by solving the following optimization problem.
\begin{align}
\arg\min_x\frac{1}{2}\|y-|\mathbf{A}x|\|^2_2+R(x),
\end{align}
where $R(x)$ is the cost function implicitly minimized by BM3D. 
%This data fidelity term, $\|y-|\mathbf{A}x|\|^2_2$, corresponds to the negative log-likelihood of $z$ under the model $y=|z|+w$ where $w\sim N(0,\sigma_w^2)$. 
Unlike the aforementioned methods, ADMM supports generic measurement matrices, including Fourier measurements.

\subsection{Neural Networks for Linear Inverse Problems}
Deep learning has recently disrupted computational imaging. %, as it has done to many other fields. 
Through the use of elaborate learned priors, deep learning methods have competed with and sometimes surpassed the performance of plug-and-play priors, while running significantly faster (when implemented on a GPU). Two prominent examples include SRCNN for superresolution \cite{SRCNN} and DnCNN for denoising, superresolution, and the removal of JPEG artifacts \cite{DnCNN}.

In addition, a few works have blended plug-and-play algorithms with neural networks \cite{OneNet, LDAMP, diamond2017unrolled}, often using algorithm unfolding/unrolling \cite{LISTA}. In doing so, these works are able to incorporate powerful learned priors, while still leveraging the flexibility and interpretability that comes from using a well-defined algorithm (as opposed to a black-box neural net).

\subsection{Neural Networks for PR}
In the last few years, researchers have raced to apply deep learning to solve the PR problem \cite{PtychNet, KaushikPR, rivenson2017phase}. So far, each of the proposed methods has been designed for a specific PR application, either Ptychography \cite{PtychNet, KaushikPR} or Holography \cite{rivenson2017phase}. The Ptychography neural networks learn to combine a stack of low-resolution band-pass-filtered images to form a high resolution image. The holographic neural network learns how to remove the twin image component \cite{goodman2005introduction} from a hologram. Because each of these networks learns an application-specific mapping, they do not generalize to new PR problems. In fact, even changing the resolution or noise level requires completely retraining these neural networks.

Our work takes a different tack. Rather than setting up a neural network to solve a specific PR problem, we use a neural network as a regularizer within an optimization framework. This technique makes our network applicable to numerous PR problems.% and lends it some interpretability.

%Can potentially deal with complex fields $x$, either by explicitly reconstructing \cite{} or ignoring \cite{}.

\section{PR via Regularization by Denoising}
In this section we first show how the Regularization by Denoising (RED) \cite{RED} framework can be adapted to solve the PR problem. We call this adaption prRED. Later, we combine prRED with the state-of-the-art DnCNN neural network \cite{DnCNN} to form prDeep.

\subsection{RED}\label{ssec:RED}
RED is an algorithmic approach to solving imaging inverse-problems that was recently proposed by Romano, Elad, and Milanfar. % \cite{RED}. 
Like many of the plug-and-play techniques described in Section \ref{ssec:pandp}, RED can incorporate an arbitrary denoiser to regularize an arbitrary imaging inverse-problem. However, whereas the other methods use a denoiser to minimize some {\em implicit} cost function, RED uses a denoiser to setup and then minimize an {\em explicit} cost function.

In particular, the RED framework defines the regularizer as
\begin{align}\label{eqn:REDreg}
R(x)=\frac{\lambda}{2} x^T(x-D(x)),
\end{align}
where $D(x)$ is an arbitrary denoiser. Note that this regularizer serves two roles. First, it penalizes the residual difference between $x$ and its denoised self; when $x-D(x)$ is large, $R(x)$ will tend to be large. Second, it penalizes correlations between $x$ and the residual. This serves to prevent $D(x)$ from removing structure from $x$; if $D(x)$ removes structure from $x$, then this structure will show up in the residual, which will be correlated with $x$. In effect, the RED regularizer encourages the residual to look like additive white Gaussian noise \cite{RED}. 

%As shown in the ADMM section of \cite{RED}, 
When $D(x)$ satisfies homogeniety and passivity conditions (see Section 3.1 of \cite{RED}) the proximal mapping of the RED regularization \eqref{eqn:REDreg} can be implemented recursively as follows\footnote{As these properties do not hold exactly in practice, \eqref{eqn:REDprox} should be considered only an approximation.}
\begin{align}\label{eqn:REDprox}
\text{prox}_R(z)&=\arg\min_x \frac{1}{2}\|x-z\|^2_2+\frac{\lambda}{2} x^T(x-D(x))\nonumber\\
&= v_\infty,\nonumber
\end{align}
where $v_j=\frac{1}{1+\lambda}(v_{j-1}+\lambda D(v_{j-1}))~\forall j>0$ and $v_0=z$.
In practice, the iterations must be terminated after a finite number $j$.
However, experiments suggest that $j=1$, which corresponds to calling the denoiser only once per use of the proximal mapping function, leads to good performance.

\subsection{prRED}

To apply RED to PR, we construct a cost function of the form
\begin{align}
f(x)+R(x),
\end{align}
where $R(x)$ is the RED regularization from \eqref{eqn:REDreg} and $f(x)$ is a
data-fidelity term that encourages $\mathbf{A}x$ to match the phaseless measurements $y$.

From a Bayesian perspective, the data-fidelity term $f(x)$ should be proportional to the negative log-likelihood function. For instance, if $y^2=|\mathbf{A}x|^2+w$ with $w\sim CN(0,\sigma_w^2\mathbf{I})$, then the negative log-likelihood function would be $-\log p(y|x)\propto \|y^2-|\mathbf{A}x|^2\|^2$.

The Bayesian perspective suggests that, when dealing with Poisson noise, which is the focus in this paper, one should use the Poisson log-likelihood function.
Interestingly, we experimented with the Poisson log-likelihood function as derived in \cite{TWF} and found that it  performed slightly worse than the amplitude loss function $f(x)=\frac{1}{2}\|y-|\mathbf{A}x|\|^2_2$. This surprising behavior was also noted in \cite{PtychRobustness}. %There, they also demonstrate that the gradient of the amplitude-loss function is very similar to that of the Poisson log-likelihood. 

In any case, a variety of data-fidelity terms can be used to solve the PR problem. 
%Zhang et al. \cite{reshapedWF}~note that amplitude loss function $\frac{1}{2}\|y-|\mathbf{A}x|\|^2_2$ works well, which is what we adopt here. 
% Like \cite{reshapedWF}, 
We  adopt the amplitude loss $\frac{1}{2}\|y-|\mathbf{A}x|\|^2_2$, which leads us to the non-convex optimization problem

\begin{align}\label{eqn:prREDcost}
\arg\min_x\frac{1}{2}\|y-|\mathbf{A}x|\|^2_2+\lambda x^T(x-D(x)).
\end{align}

Although various solvers could be used to attack \eqref{eqn:prREDcost}, we use the FASTA solver \cite{FASTA}. FASTA implements the forward-backward splitting algorithm (a.k.a.~the proximal gradient method) and incorporates adaptive step sizes for acceleration. FASTA is very simple to use; after defining the loss function, one need only provide the solver with a proximal mapping for the regularization term \eqref{eqn:REDprox} and a (sub)gradient for the data-fidelity term with respect to $z$ (for $z\triangleq\mathbf{A}x$). For our adopted data-fidelity term, a useful subgradient is 
\begin{align}\label{eqn:subGrad}
z-y\circ \frac{z}{|z|} \in \partial_z \frac{1}{2}\|y-|z|\|^2_2,
\end{align}
where $\circ$ denotes the Hadamard (i.e., elementwise) product and $\partial_zf(z)$ denotes the subdifferential of $f$ with respect to $z$.
 
\begin{figure}[t]
	\centering
	\includegraphics[width=.4\textwidth]{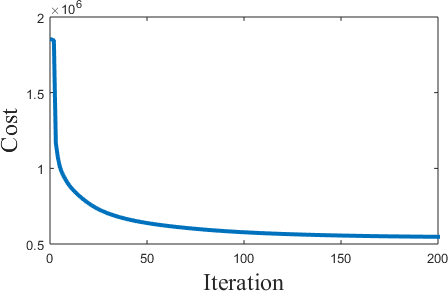}
	\caption{The cost \eqref{eqn:prREDcost} over the number of FASTA iterations. The cost decreases monotonically.}
	\label{fig:Cost}
\end{figure}

In practice, the FASTA solver converges very quickly. As an example, Figure \ref{fig:Cost} shows a typical cost \eqref{eqn:prREDcost} per iteration trajectory for the FASTA solver. There it can be seen that the cost drops monotonically and converges after about 200 iterations.

\subsection{prDeep}
%\subsubsection{DnCNN Denoiser}

\begin{figure}[t]
	\centering
	\includegraphics[width=.45\textwidth]{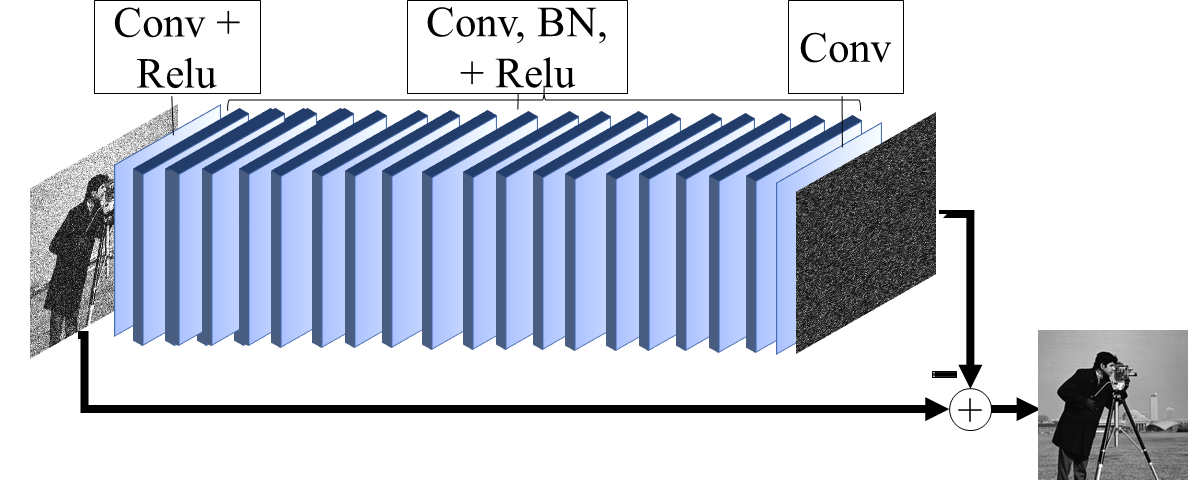}
	\caption{Network architecture of the DnCNN denoiser. Note that it takes advantage of residual learning.}
	\label{fig:DnCNN}
\end{figure}
The prRED framework can incorporate nearly any denoising algorithm. We call the special case of prRED with the DnCNN denoiser ``prDeep''.

DnCNN \cite{DnCNN} is a state-of-the-art denoiser for removing additive white Gaussian noise from natural images. DnCNN consists of 16 to 20 convolutional layers of size $3\times3$ (we used 20). Sandwiched between these layers are ReLU \cite{Relu} and batch-normalization \cite{batchnorm} operations. DnCNN is trained using residual learning \cite{residuallearning}.

In practice, DnCNN noticeably outperforms the popular BM3D algorithm. Moreover, thanks to parallelization and GPU computing, it runs hundreds of times faster than BM3D.

We trained four DnCNN networks at different noise levels. To train, we loosely followed the procedure outlined in \cite{DnCNN}. In particular, we trained with 300\,000 overlapping patches drawn from $400$ images in the Berkeley Segmentation Dataset \cite{BSDDataset}. For each image patch, we added additive white Gaussian noise with a standard deviation of either $60$, $40$, $20$, or $10$. where our images had a dynamic range of $[0,255]$. We then setup DnCNN to recover the noise-free image.
We used the mean-squared error between the noise-free ground truth image and our denoised reconstructions as the cost function. We trained the network with stochastic gradient descent and the ADAM optimizer \cite{ADAMopt} with a batch size of 256. Our training rate was $0.001$, which we dropped to $0.0001$ and then $0.00001$ when the validation error stopped improving. Training took just over 3 hours per noise level on an Nvidia Pascal Titan X.

%\subsection{Understanding prDeep}
%\textcolor{red}{Is there any theory we can say about prRED/prDeep? If we can think of anything this could be its own section.}
%prDeep can be interpreted as recurrent neural network.
%\input{training.tex}
\section{Experimental Results}\label{sec:Results}
\begin{figure}[t]
\centering
\subfigure[Barbara]{\includegraphics[width=.14\textwidth]{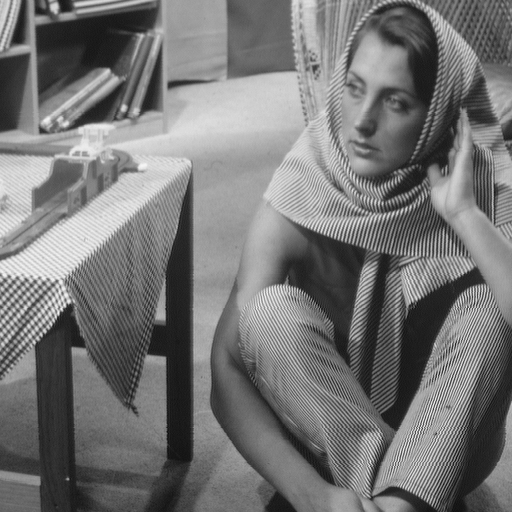}}
\subfigure[Boat]{\includegraphics[width=.14\textwidth]{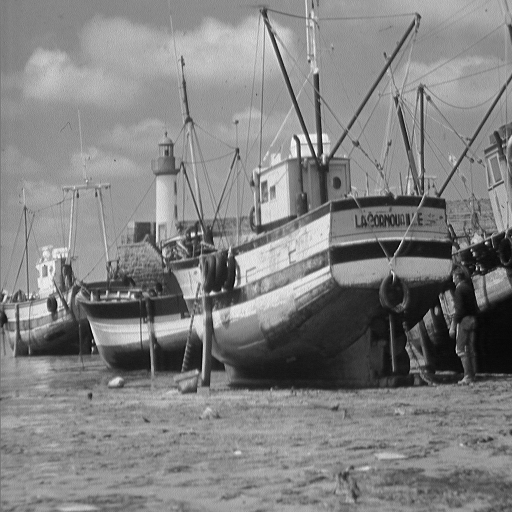}}
\subfigure[Couple]{\includegraphics[width=.14\textwidth]{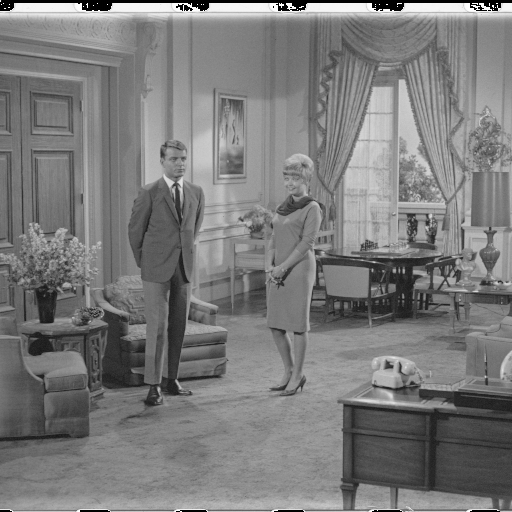}}
\subfigure[Peppers]{\includegraphics[width=.14\textwidth]{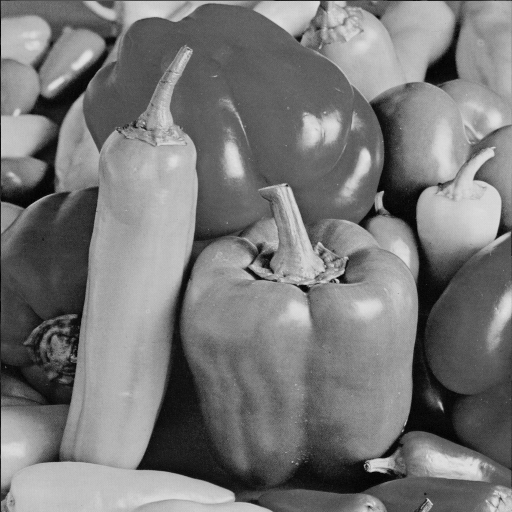}}
\subfigure[Cameraman]{\includegraphics[width=.14\textwidth]{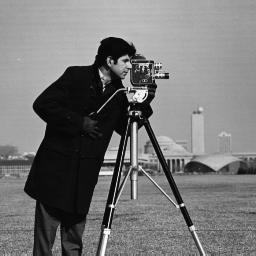}}
\subfigure[Bridge]{\includegraphics[width=.14\textwidth]{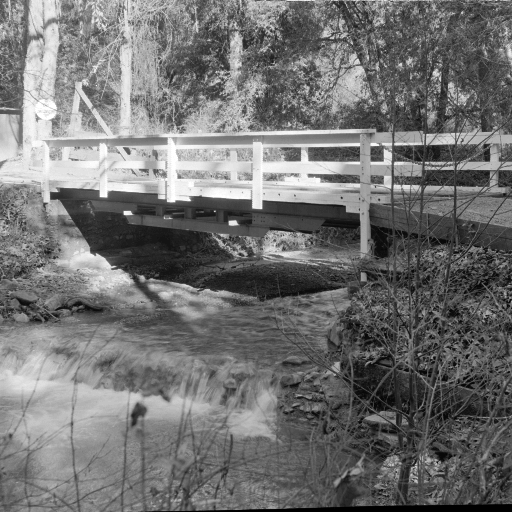}}
\caption{The six ``natural'' test images. They follow a distribution very similar to the training data.}
\label{fig:NatTestImages}
\end{figure}

\begin{figure}[t]
\centering
\subfigure[E.~Coli]{\includegraphics[width=.14\textwidth]{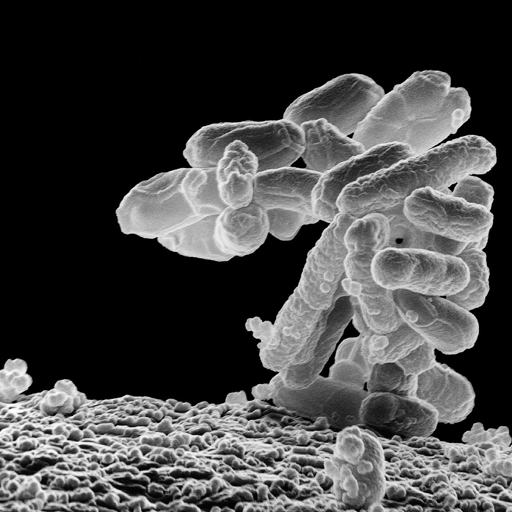}}
\subfigure[Yeast]{\includegraphics[width=.14\textwidth]{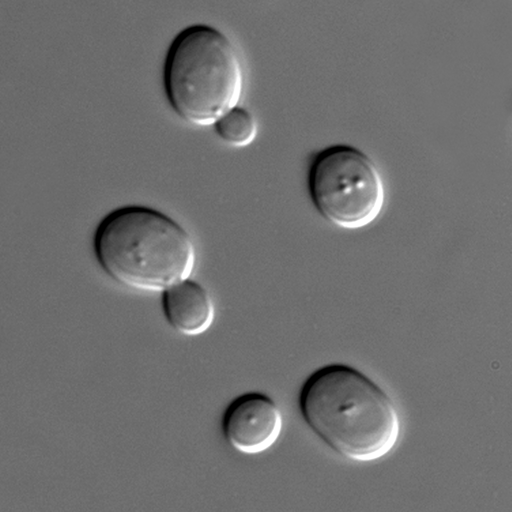}}
\subfigure[Pollen]{\includegraphics[width=.14\textwidth]{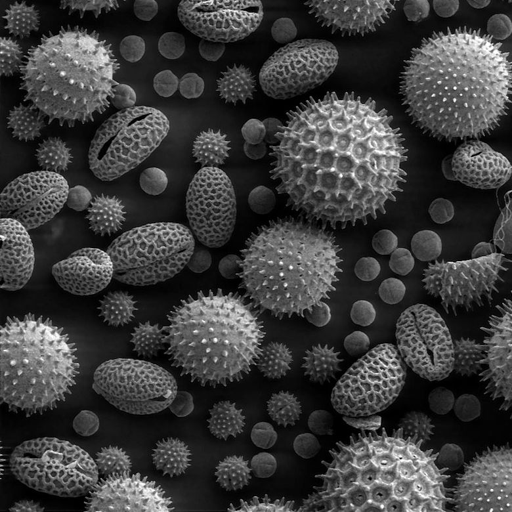}}
\subfigure[Tadpole Galaxy]{\includegraphics[width=.14\textwidth]{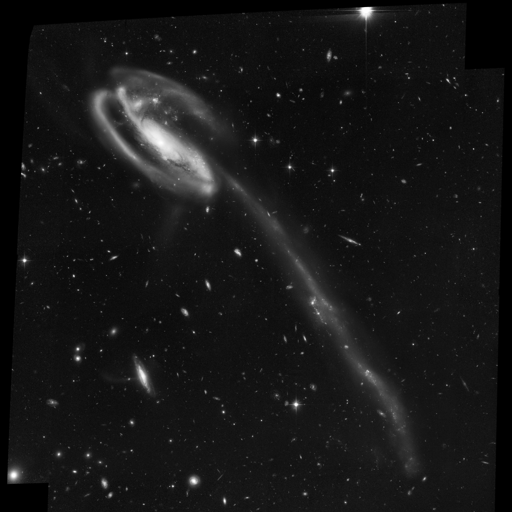}}
\subfigure[Pillars of Creation]{\includegraphics[width=.14\textwidth]{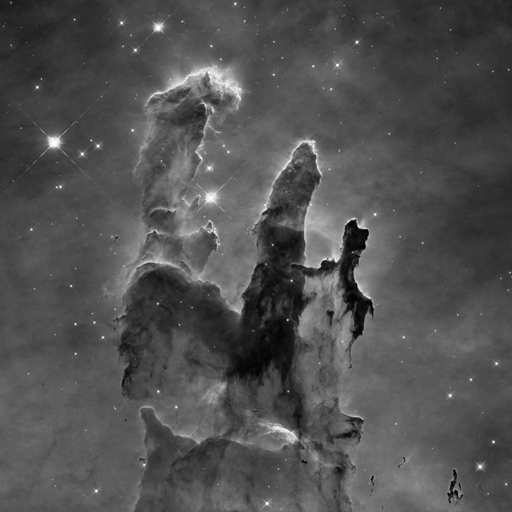}}
\subfigure[Butterfly Nebula]{\includegraphics[width=.14\textwidth]{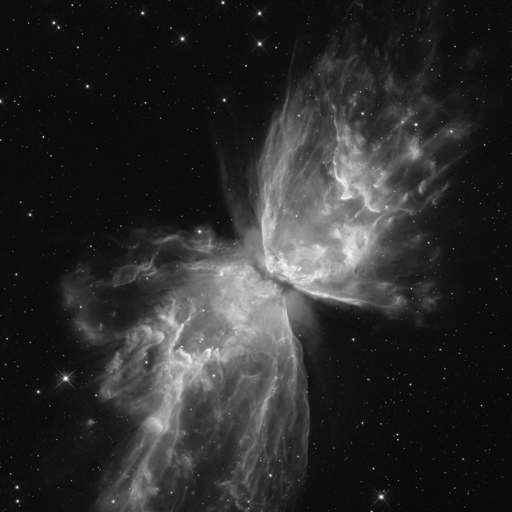}}
\caption{The six ``unnatural'' test images. They follow a distribution distinct from the training data. Images are from Wikipedia and follow public domain licenses.}
\label{fig:UnnatTestImages}
\end{figure}

In this section we compare prDeep to several other PR algorithms on simulated data with varying amounts of Poisson noise. We test the algorithms with both coded diffraction pattern (CDP) and Fourier measurements. In both sets of tests, we sample and reconstruct 6 ``natural'' and 6 ``unnatural'' (real and nonnegative) test images, which are presented in Figures \ref{fig:NatTestImages} and \ref{fig:UnnatTestImages}. %Note that none of the test images are symmetric; symmetric images are far more challenging to reconstruct. 

\subsection{Experimental Setup}
\paragraph{Competing Algorithms.}
We compare prDeep against Hybrid Input-Output (HIO) \cite{HIO}, Oversampling Smoothness (OSS) \cite{OSS}, Wirtinger Flow (WF) \cite{candes2015phase}, DOLPHIn \cite{dolphin}, SPAR \cite{SPAR1,SPAR2}, and BM3D-prGAMP \cite{bm3dprgamp}. We also compare with Plug-and-Play ADMM \cite{P&PADMM,Proximal} using both the BM3D and DnCNN denoisers. 
HIO and WF are baseline algorithms designed for Fourier and CDP measurements, respectively.
%HIO is a simple alternating projection method developed by Fienup over 30 years ago. It performs remarkably well and remains widely used in practice today. 
OSS is an alternating projection algorithm designed for noisy Fourier measurements. It imposes a smoothness constraint to regions outside of the target's support.
%\textcolor{red}{I might want to cut OSS. Its performance was very inconsistent}
%WF is stochastic gradient descent like algorithm that comes with nice theoretical guarantees.
DOLPHIn is an iterative PR algorithm designed to reconstruct images from noisy CDP measurements. It imposes a sparsity constraint with respect to a learned dictionary.
SPAR, BM3D-prGAMP, and Plug-and-Play ADMM were described in Section \ref{ssec:P&PforPR}.% Both algorithms were designed for CDP measurements.
%We do not compare against the commonly used ER-HIO \cite{} and NR-HIO \cite{} algorithms, because in \cite{OSS} it was demonstrated that OSS soundly outperforms them.

%Unfortunately, we are not able to compare against Plug-and-Play ADMM. While \cite{Proximal} demonstrated that this algorithm could be adapted to PR, the suggested modifications are nontrivial and their implementation is not publicly available at this time.

%In addition to the aforementioned algorithms, tn order to separate what portion of prDeep's performance comes from the RED framework versus what comes from DnCNN, we also test against FASTA \cite{} with simple a call to DnCNN as the proximal mapping. \textcolor{red}{Need to test this}. Can add this in a revision

\begin{table*}[t]
  \centering
  \caption{PSNRs and run times (sec) of $128\times128$ reconstructions with four intensity-only CDP measurements and varying amounts of Poisson noise.}
    \begin{tabular}{lrrrrrrrrr}
    \toprule
      & \multicolumn{3}{c}{$\alpha=9$} & \multicolumn{3}{c}{$\alpha=27$} & \multicolumn{3}{c}{$\alpha=81$} \\
     \cmidrule{2-10}
     & \multicolumn{1}{l}{\makecell{PSNR\\Natural}} & \multicolumn{1}{l}{\makecell{PSNR\\Unnatural}} & \multicolumn{1}{l}{Time} & \multicolumn{1}{l}{\makecell{PSNR\\Natural}} & \multicolumn{1}{l}{\makecell{PSNR\\Unnatural}} & \multicolumn{1}{l}{Time} & \multicolumn{1}{l}{\makecell{PSNR\\Natural}} & \multicolumn{1}{l}{\makecell{PSNR\\Unnatural}} & \multicolumn{1}{l}{Time} \\
    \midrule
    HIO   & 36.1  & 36.0  & 4.2   & 26.1  & 26.0  & 5.9   & 14.8  & 16.9  & 5.9 \\
    WF    & 34.3  & 34.2  & 0.8   & 24.7  & 24.0  & 0.9   & 13.2  & 13.0  & 6.7 \\
    DOLPHIn & 31.0  & 28.8  & 1.1   & 27.9  & 26.9  & 1.1   & 17.9  & 20.8  & 1.1 \\
    SPAR  & 34.4  & 36.0  & 15.7  & 29.4  & 31.1  & 15.7  & \textbf{24.5} & 26.4  & 15.7 \\
    BM3D-prGAMP & 34.1  & 35.4  & 17.2  & 31.4  & 32.2  & 17.7  & 22.5  & 22.3  & 23.1 \\
    BM3D-ADMM & 38.5  & 39.0  & 51.6  & 31.9  & 33.4  & 40.0  & 24.4  & \textbf{26.9} & 37.7 \\
    DnCNN-ADMM & \textbf{39.9} & \textbf{40.6} & 17.1  & 32.6  & 33.4  & 9.8   & 21.6  & 22.7  & 7.9 \\
    prDeep & 39.1  & 40.1  & 25.4  & \textbf{32.8} & \textbf{34.0} & 23.2  & 20.9  & 24.5  & 23.1 \\
    \bottomrule
    \end{tabular}%

  \label{tab:diffraction}%
\end{table*}%

\begin{table*}[t]
  \centering
  \caption{PSNRs and run times (sec) of $128\times128$ reconstructions with $4\times$ over-sampled intensity-only Fourier measurements and varying amounts of Poisson noise. }
    \begin{tabular}{lrrrrrrrrr}
    \toprule
      & \multicolumn{3}{c}{$\alpha=2$} & \multicolumn{3}{c}{$\alpha=3$} & \multicolumn{3}{c}{$\alpha=4$} \\
     \cmidrule{2-10}
     & \multicolumn{1}{l}{\makecell{PSNR\\Natural}} & \multicolumn{1}{l}{\makecell{PSNR\\Unnatural}} & \multicolumn{1}{l}{Time} & \multicolumn{1}{l}{\makecell{PSNR\\Natural}} & \multicolumn{1}{l}{\makecell{PSNR\\Unnatural}} & \multicolumn{1}{l}{Time} & \multicolumn{1}{l}{\makecell{PSNR\\Natural}} & \multicolumn{1}{l}{\makecell{PSNR\\Unnatural}} & \multicolumn{1}{l}{Time} \\
    \midrule
    HIO   & 22.2  & 20.8  & 10.6  & 19.9  & 17.9  & 10.4  & 17.9  & 15.5  & 10.7 \\
    WF    & 15.2  & 18.8  & 6.7   & 15.1  & 18.5  & 6.6   & 15.2  & 18.5  & 6.7 \\
    OSS   & 18.6  & 23.8  & 18.3  & 18.4  & 23.1  & 18.6  & 18.6  & 22.7  & 18.1 \\
    SPAR  & 22.0  & 23.9  & 79.1  & 19.6  & 22.6  & 77.0  & 19.4  & 20.5  & 83.0 \\
    BM3D-prGAMP & 24.5  & 26.0  & 84.3  & 23.2  & 24.4  & 82.8  & 21.0  & 22.7  & 84.4 \\
    BM3D-ADMM & 27.5  & 28.3  & 155.7 & 24.5  & 25.0  & 156.8 & 21.8  & 23.1  & 156.8 \\
    DnCNN-ADMM & \textbf{29.3} & \textbf{31.3} & 65.0  & 26.0  & 25.8  & 62.4  & 22.0  & 23.4  & 64.0 \\
    prDeep & 28.4  & 30.6  & 105.1 & \textbf{28.5} & \textbf{26.4} & 105.0 & \textbf{26.4} & \textbf{25.1} & 107.1 \\
    \bottomrule
    \end{tabular}%
    
  \label{tab:Fourier}%
\end{table*}%

\paragraph{Implementation.} All algorithms were tested using Matlab 2017a on a desktop PC with an Intel 6800K CPU and an Nvidia Pascal Titan X GPU. Dolphin, SPAR, and BM3D-prGAMP used their respective authors' implementations. We created our own version of Plug and Play ADMM based off of code original developed in \cite{chan2017plug}. 
We use FASTA to solve the nonlinear least squares problem at each iteration of the algorithm. 
prDeep and DnCNN-ADMM used a MatConvNet \cite{vedaldi15matconvnet} implementation of DnCNN. A public implementations of prDeep is available at \url{https://github.com/ricedsp/prDeep}.

\paragraph{Measurement and Noise Model.}
Model mismatch and Poisson shot noise are the dominant sources of noise in many PR applications \cite{PtychRobustness}. 
In this paper, we focus on shot noise, which we approximate as
%To form the Poisson measurement vector $y^2$, we generate an independent Gaussian random vector $w$ with component-wise variance proportional to $|z|^2$, and then added it to $z$. That is
\begin{align}
y^2=|z|^2+w\text{ with }w\sim N(0,\alpha^2 \text{Diag}(|z|^2)),\nonumber
\end{align}
where $z=\mathbf{A}x$ with known $\mathbf{A}$, and where $\text{Diag}(|z|^2)$ is a diagonal matrix with diagonal elements $|z|^2$.

% In this paper we focus on the shot noise and model our measurements as follows
% \begin{align}
%     y^2=\text{Poisson}(|z|^2),\nonumber
% \end{align}
% where $z=\mathbf{A}x$ and $\mathbf{A}$ is known exactly. 

%One could potentially rescale $\mathbf{A}$ or $x$ in order to control $|z|^2$, and thus the difficulty of the PR problem. However the resulting 
%To form the Poisson measurement vector $y^2$, we generate an independent Gaussian random vector $w$ with component-wise variance proportional to $|z|^2$, i.e., $w\sim N(0,\text{diag}(|z|^2))$. For large values of $|z|$, the random vector $y^2=|z|^2+w$ is approximately $\text{Poisson}(|z|^2)$.

Some algebra and the central limit theorem can be used to show that $y^2/\alpha^2\sim \text{Poisson}((|z|/\alpha)^2)$. In effect, $y^2$ is a rescaled Poisson random variable. The term $\alpha$ controls the variance of the random variable and thus the effective signal-to-noise ratio in our problem.

\paragraph{Parameter Tuning.}
HIO was run for $1000$ iterations.
WF was run for $2000$ iterations. %, or until its residual reached $1e-6$.
%We disabled the Phase unwrapping and filtering options in SPAR because our images were real-valued.
BM3D-prGAMP was run for $50$ iterations.
prDeep was run for $200$ iterations four times; once for each of the denoisers networks (trained at standard deviations $60$, $40$, $20$, and $10$). The result from reconstructing with the first denoiser was used to warm-start reconstructing with the second, the second warm-started the third, etc. %for noise with standard deviations $60$, $40$, $20$, and $10$.
Plug and Play ADMM was similarly run for $50$ iterations four times.  ADMM converged faster than prDeep and did not benefit from additional iterations.

SPAR reconstructed $x$ using $y^2/\alpha^2$ rather than $y$, as its loss function expects Poisson distributed random variables.
prDeep's parameter $\lambda$, which determines the amount of regularization, was set to $\bar{\sigma}_w$ when dealing with Fourier measurements and $0.1\bar{\sigma}_w$ when dealing with CDP measurements, where $\bar{\sigma}_w^2$ denotes the sample variance of the noise.
The parameter in ADMM analogues to $\lambda$ was set to $.2\bar{\sigma}_w$ when dealing with Fourier measurements and $0.02\bar{\sigma}_w$ when dealing with CDP measurements.
The algorithms otherwise used their default parameters.

\paragraph{Initialization.}
With oversampled CDP measurements of real-valued signals, none of the algorithms were particularly sensitive to initialization; initializing with a vector of ones worked sufficiently well.
In contrast, with Fourier measurements, the algorithms were very sensitive to initialization. We experimented with various spectral initializers, but found they were ineffective with (noisy and only $4\times$ oversampled) Fourier measurements. Instead, we first ran the HIO algorithm (for $50$ iterations) $50$ times, from random initializations, to form 50 estimates of the signal: $\widehat{x}_1$, $\widehat{x}_2$, ... $\widehat{x}_{50}$. We then used the reconstruction $\widehat{x}_i$ with the lowest residual ($\|y-|\mathbf{A}\widehat{x}_i|\|_2$) as an initialization for HIO. HIO was then run for $1000$ iterations, and the result was used to initialize the other algorithms. This process was repeated three times and the reconstruction with the smallest residual was used as the final estimate. The reported computation times (for Fourier measurements) include the time required to initialize the algorithms and run them three times.%\cite{Proximal} also initialized their Plug-and-Play ADMM algorithm with the solution from HIO.

%When testing oversampled Fourier measurements, we provided the algorithms an oracle estimate of the support. \textcolor{red}{Need to describe how the support was encoded into the rectangular measurement matrix used by prDeep}

%As in the Plug and Play ADMM work \cite{Proximal} we initialized prDeep with the solution from HIO. 

%\textcolor{red}{Can cite Laura Waller's paper to justify this approximation \url{https://www.osapublishing.org/DirectPDFAccess/6D75F398-D3DF-4D69-9DF06ECBF8EE61B3_333747/oe-23-26-33214.pdf?da=1&id=333747&seq=0&mobile=no}}
%One could also use \texttt{poissrnd} to generate Poisson distributed random variables, but controllinf for the various dynamic ranges this produces can become a real challenge.

\subsection{Simulated Coded Diffraction Measurements}

\begin{figure*}[t]
\centering
\subfigure[HIO (21 sec)]{\includegraphics[width=.22\textwidth]{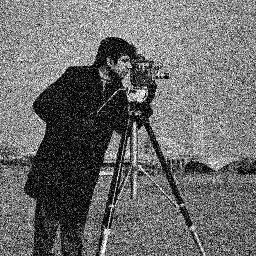}}
\subfigure[WF (24 sec)]{\includegraphics[width=.22\textwidth]{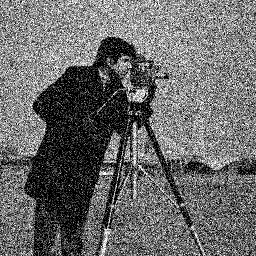}}
\subfigure[DOLPHIn (3 sec)]{\includegraphics[width=.22\textwidth]{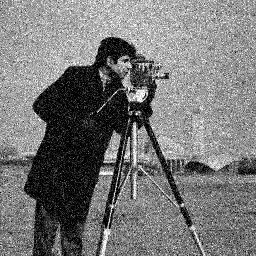}}
\subfigure[SPAR (58 sec)]{\includegraphics[width=.22\textwidth]{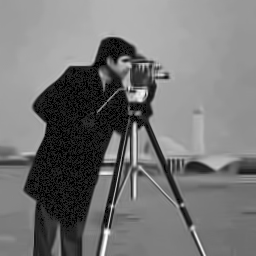}}
\subfigure[BM3D-prGAMP (91 sec)]{\includegraphics[width=.22\textwidth]{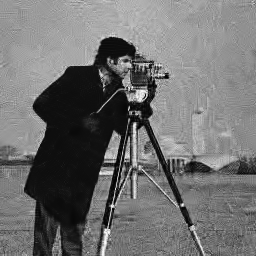}}
\subfigure[BM3D-ADMM (153 sec)]{\includegraphics[width=.22\textwidth]{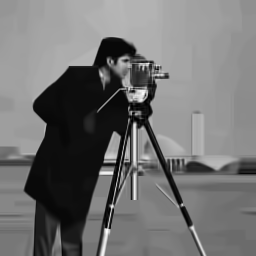}}
\subfigure[DnCNN-ADMM (25 sec)]{\includegraphics[width=.22\textwidth]{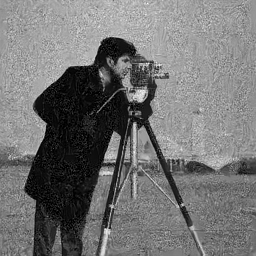}}
\subfigure[prDeep (72 sec)]{\includegraphics[width=.22\textwidth]{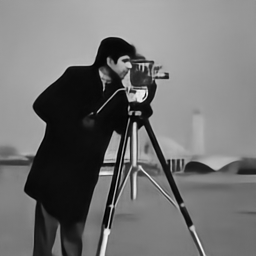}}
\caption{Reconstructions of $256\times 256$ image from four noisy intensity-only CDP measurements ($\alpha=81$) with several PR algorithms. All five plug-and-play methods provide similar reconstructions.}
\label{fig:CDI_visual}
\end{figure*}

We first test the algorithms with CDP intensity-only measurements.
CDP is a measurement model proposed in \cite{candesCodedDiffraction} that uses a spatial light modulator (SLM) to spread a target's frequency information and make it easier to reconstruct. 
Under a CDP measurement model, the target is illuminated by a coherent source and then has its phase immediately modulated by a known random pattern using an SLM. The complex field then undergoes far-field Fraunhofer diffraction, which can be modeled by a 2D Fourier transform, before its intensity is recorded by a standard camera. Multiple measurements, with different random SLM patterns, are recorded. In this work, we model the capture of four measurements using a phase-only SLM. Mathematically, our measurement operator is as follows
\begin{align}
    \mathbf{A}=\left[
        \begin{array}{ll}
        \mathbf{F}\mathbf{D_1}\\
        \mathbf{F}\mathbf{D_2}\\
        \mathbf{F}\mathbf{D_3}\\
        \mathbf{F}\mathbf{D_4}
        \end{array}
        \right],
\end{align}
where $\mathbf{F}$ represents the 2D Fourier transform and $\mathbf{D}_1$, $\mathbf{D}_2$, ... are diagonal matrices with nonzero elements drawn uniformly from the unit circle in the complex plane.

In Table \ref{tab:diffraction} we compare the performance of the various PR algorithms. 
We do not include a comparison with OSS as it is setup specifically for Fourier measurements.
%\textcolor{red}{It might be best to drop OSS entirely. Even with Fourier measurements the recovery accuracies were horrible.}
We report recovery accuracy in terms of mean peak-signal-to-noise ratio (PSNR) across two sets of test images.\footnotetext{${\rm PSNR} = 10 \log_{10}(\frac{255^2}{{\rm mean}((\hat{x}-x_o)^2)})$ when the pixel range is 0 to 255.}
We report run time in seconds. 
%We also report the median effective SNR of the measurement $y$, defined as $10\log_{10}(\||\mathbf{A}x|^2\|_2^2/\|w\|_2^2)$.

Table \ref{tab:diffraction} demonstrates that, when dealing with CDP measurements at low SNRs (large $\alpha$), all five plug-and-play methods produce similar reconstructions.% At high SNRs (small $\alpha$) prDeep performs slightly worse; its current configuration produces slightly oversmoothed reconstructions.
%prDeep and BM3D-prGAMP are faster than SPAR.

In Figure \ref{fig:CDI_visual}, we visually compare the reconstructions of a $256\times256$ image using the algorithms under test. With CDP measurements, all of the plug-and-play algorithms do a good job reconstructing the signal from noisy measurements.% BM3D-prGAMP is not included as it consistently diverged with noisy high dimensional measurements.

%See "Computational super-resolution PR from multiple phase-coded diffraction patterns: simulation study and experiments". Their data might be public too.

\subsection{Simulated Fourier Measurements}
\begin{figure*}[t]
\centering
\subfigure[HIO (40 sec)]{\includegraphics[width=.22\textwidth]{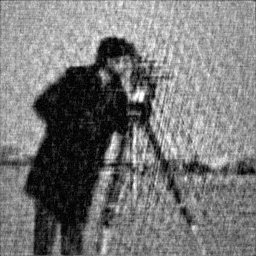}}
\subfigure[WF (63 sec)]{\includegraphics[width=.22\textwidth]{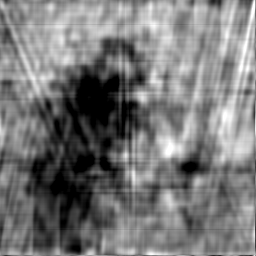}}
\subfigure[OSS (65 sec)]{\includegraphics[width=.22\textwidth]{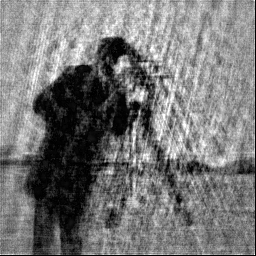}}
\subfigure[SPAR (294 sec)]{\includegraphics[width=.22\textwidth]{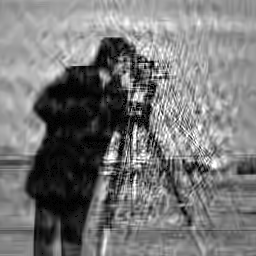}}
\subfigure[BM3D-prGAMP (306 sec)]{\includegraphics[width=.22\textwidth]{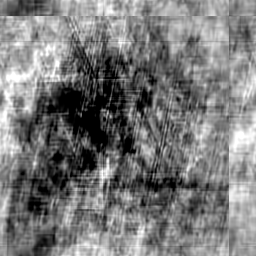}}
\subfigure[BM3D-ADMM (576 sec)]{\includegraphics[width=.22\textwidth]{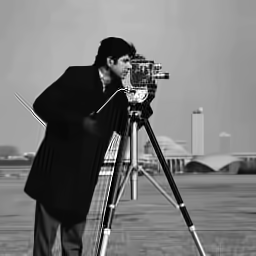}}
\subfigure[DnCNN-ADMM (426 sec)]{\includegraphics[width=.22\textwidth]{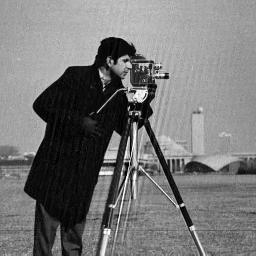}}
\subfigure[prDeep (345 sec)]{\includegraphics[width=.22\textwidth]{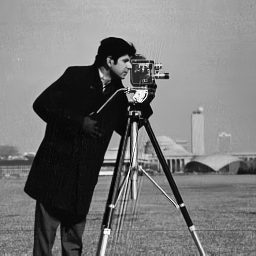}}
\caption{Reconstructions of $256\times 256$ image from noisy $4\times$ oversampled intensity-only Fourier measurements ($\alpha=3$) with several PR algorithms. Plug and Play ADMM and prDeep provide the best reconstructions.}
\label{fig:Fourier_visual}
\end{figure*}

Fourier measurements of real-valued signals are prevalent in many real-world applications. 
Many PR applications exploit the Fourier-transform property $\mathbf{F}(x\star x)=|\mathbf{F}x|^2$, where $\star$ denotes correlation. This implies that in applications where one can measure or estimate the autocorrelation function of an object one can also measure the modulus squared of its Fourier transform. This allows PR algorithms to reconstruct the object.
%In correlography one has a sequence, that with infinite measurements, converges to the autocorrelation function. Your not directly measuring the autocorrelation, but "estimating" is a little weak.

This relationship has been used in multiple contexts. In astronomical imaging, this relationship has been used to image through turbulent atmosphere \cite{fienup1987phase}. In laser-illuminated imaging, this relationship has been used to reconstruct diffuse objects without speckle noise \cite{correlography}. More recently, this relationship has been used to image through random scattering media, such as biological tissue \cite{katz2014non}. In all of these applications, one reconstructs the { real-valued} intensity distribution of the object.

In these tests we oversampled the spectrum by $4\times$. That is, we first placed $128\times128$ images at the center of a $256\times256$ square and then took the 2D Fourier transform. We assumed that the support, i.e.,~the location of the image within the $256\times256$ grid, was known a priori.

Table \ref{tab:Fourier} compares the reconstruction accuracies and recovery times of several PR algorithms.\footnote{Our results account for the translation and reflection ambiguities associated with Fourier measurements} %The recovery times of SPAR, BM3D-prGAMP, BM3D-ADMM, DnCNN-ADMM, and prDeep include the time taken to initialize with HIO and run the algorithm three times. 
We do not include results for DOLPHIn, as it completely failed with Fourier measurements. 
Note that, at a given noise level, the reconstructions from Fourier measurements are far less accurate than their CDP counterpart. Table \ref{tab:Fourier} demonstrates that with Fourier measurements and large amounts of noise prDeep is superior to existing PR algorithms.% Its computation time is competitive with other methods and is dominated by the cost of initializing with HIO.

In Figure \ref{fig:Fourier_visual}, we again visually compare the reconstructions from the  algorithms under test, this time with Fourier measurements. In this regime prDeep produces fewer artifacts than competing methods.

\section{Conclusions and Future Work}

In this paper, we have extended and applied the Regularization by Denoising (RED) framework to the problem of PR. 
Our new algorithm, prDeep, is exceptionally robust to noise thanks to the use of the DnCNN image denoising neural network. 
As we demonstrated in our experiments, prDeep is also able to handle a wide range of measurement matrices, from intensity-only coded diffraction patterns to Fourier measurements.

By integrating a neural network into a traditional optimization algorithm, prDeep inherits the strengths of both optimization and deep-learning. Like other optimization-based algorithms, prDeep is flexible and can be applied to PR problems with different measurement models, noise levels, etc., without having to undergo costly retraining. Like other deep-learning-based techniques, for any given problem prDeep can take advantage of powerful learned priors and outperform traditional, hand-designed methods.

%Unlike many PR algorithms, 
prDeep is not a purely academic PR algorithm; it can handle Fourier measurements of real-valued signals, a measurement model that plays a key role in many imaging applications. However, prDeep does have a major limitation; it is presently restricted to amplitude-only targets.
Extending prDeep to handle complex-valued targets is a promising and important direction for future research. 

\subsection*{Acknowledgements}
%Thanks to XXX and YYY for insightful discussions and making coffee.
Phil Schniter was supported by NSF grants CCF-1527162 and CCF-1716388. 
Richard Baraniuk, Ashok Veeraraghavan, and Chris Metzler were supported by the DOD Vannevar Bush Faculty Fellowship N00014-18-1-2047, NSF Career– IIS-1652633, and the NSF GRF program, respectively. 
They were also supported by NSF grant CCF-1527501, ARO grant W911NF-15-1-0316, AFOSR grant FA9550-14-1-0088, ONR grant N00014-17-1-2551, DARPA REVEAL grant HR0011-16-C-0028, ARO grant Supp-W911NF-12-1-0407, and an ONR BRC grant for Randomized Numerical Linear Algebra.

\FloatBarrier
\bibliography{egbib}
\bibliographystyle{icml2018}

\end{document}